\documentclass[10pt,twocolumn,letterpaper]{article}

\pdfoutput=1
\usepackage{cvpr}               

\usepackage[accsupp]{axessibility}
\usepackage{graphicx}
\usepackage{capt-of}
\usepackage{xcolor}
\usepackage{amsmath,amssymb,amsfonts,dsfont,pifont,bm,bbm,mathrsfs,mathtools,nicefrac}
\usepackage{algorithm,algpseudocode,listings}
\usepackage{booktabs,multirow,adjustbox,diagbox,threeparttable}
\definecolor{citeblue}{RGB}{48,111,186}
\usepackage[pagebackref,breaklinks,colorlinks,citecolor=citeblue,bookmarks=false]{hyperref}
\usepackage[capitalize]{cleveref}  
\crefname{section}{Sec.}{Secs.}
\Crefname{section}{Section}{Sections}
\crefname{table}{Tab.}{Tabs.}
\Crefname{table}{Table}{Tables}
\crefname{figure}{Fig.}{Figs.}
\Crefname{figure}{Figure}{Figures}
\crefname{equation}{Eq.}{Eqs.}
\Crefname{equation}{Equation}{Equations}
\def\confName{CVPR}
\def\confYear{2023}

\begin{document}

\title{Taming Diffusion Models for Audio-Driven Co-Speech Gesture Generation}

\author{Lingting Zhu$^{1}$\thanks{Equal contribution.}\quad Xian Liu$^{2}$\footnotemark[1]\quad Xuanyu Liu$^{1}$\quad Rui Qian$^{2}$\quad Ziwei Liu$^{3}$\quad Lequan Yu$^{1}$\thanks{Corresponding author.}\\
$^1$The University of Hong Kong \ \  $^2$The Chinese University of Hong Kong \ \  \\$^3$S-Lab, Nanyang Technological University \\
    {\tt\small \{ltzhu99, u3008631\}@connect.hku.hk, lqyu@hku.hk,}\\
    {\tt\small \{alvinliu, qr021\}@ie.cuhk.edu.hk, }
    {\tt\small ziwei.liu@ntu.edu.sg}
}

\maketitle

\def\@fnsymbol#1{\ensuremath{\ifcase#1\or *\or \dagger\or  \ddagger\or
  \mathsection\or \mathparagraph\or \|\or **\or \dagger\dagger
  \or \ddagger\ddagger \else\@ctrerr\fi}}
\renewcommand{\thefootnote}{\fnsymbol{footnote}}
   
\begin{abstract}

Animating virtual avatars to make co-speech gestures facilitates various applications in human-machine interaction.
The existing methods mainly rely on generative adversarial networks (GANs), which typically suffer from notorious mode collapse and unstable training, thus making it difficult to learn accurate audio-gesture joint distributions.
In this work, we propose a novel diffusion-based framework, named \textbf{Diffusion Co-Speech Gesture (DiffGesture)}, to effectively capture the cross-modal audio-to-gesture associations and preserve temporal coherence for high-fidelity audio-driven co-speech gesture generation. 
Specifically, we first establish the diffusion-conditional generation process on clips of skeleton sequences and audio to enable the whole framework.
Then, a novel Diffusion Audio-Gesture Transformer is devised to better attend to the information from multiple modalities and model the long-term temporal dependency. 
Moreover, to eliminate temporal inconsistency, we propose an effective Diffusion Gesture Stabilizer with an annealed noise sampling strategy.
Benefiting from the architectural advantages of diffusion models, we further incorporate implicit classifier-free guidance to trade off between diversity and gesture quality. Extensive experiments demonstrate that DiffGesture achieves state-of-the-art performance, which renders coherent gestures with better mode coverage and stronger audio correlations.
Code is available at {\href{https://github.com/Advocate99/DiffGesture}{https://github.com/Advocate99/DiffGesture}}.
\end{abstract}
\section{Introduction}\label{sec:intro}

Making co-speech gestures is an innate human behavior in daily conversations, which helps the speakers to express their thoughts and the listeners to comprehend the meanings~\cite{cassell1999speech, mcneill2011hand, 2014Gesture}. 
Previous linguistic studies verify that such non-verbal behaviors could liven up the atmosphere and improve mutual intimacy~\cite{burgoon1990nonverbal, 1989Gesture, huang2012robot}.
Therefore, animating virtual avatars to gesticulate co-speech movements is crucial in embodied AI. 
To this end, recent researches focus on the problem of audio-driven co-speech gesture generation~\cite{ginosar2019learning, yoon2020speech, liu2022learning, li2021audio2gestures}, which synthesizes human upper body gesture sequences that are aligned to the speech audio.

Early attempts downgrade this task as a searching-and-connecting problem, where they predefine the corresponding gestures of each speech unit and stitch them together by optimizing the transitions between consecutive motions for coherent results~\cite{cassell1994animated, huang2012robot, marsella2013virtual}. 
In recent years, the compelling performance of deep neural networks has prompted data-driven approaches. 
Previous studies establish large-scale speech-gesture corpus to learn the mapping from speech audio to human skeletons in an end-to-end manner~\cite{alexanderson2020style, liu2022beat, xu2022freeform, qian2021speech, liu2022learning, li2021audio2gestures, ao2022rhythmic}. 
To attain more expressive results, Ginosar \textit{et al.}~\cite{ginosar2019learning} and Yoon \textit{et al.}~\cite{yoon2020speech} propose GAN-based methods to guarantee realism by adversarial mechanism, where the discriminator is trained to distinguish real gestures from the synthetic ones while the generator's objective is to fool the discriminator. 
However, such pipelines suffer from the inherent mode collapse and unstable training, making them difficult to capture the \textit{high-fidelity audio-conditioned} gesture distribution, resulting in dull or unreasonable poses.

\begin{figure}[t]
\centering
\includegraphics[width=1.00\columnwidth]{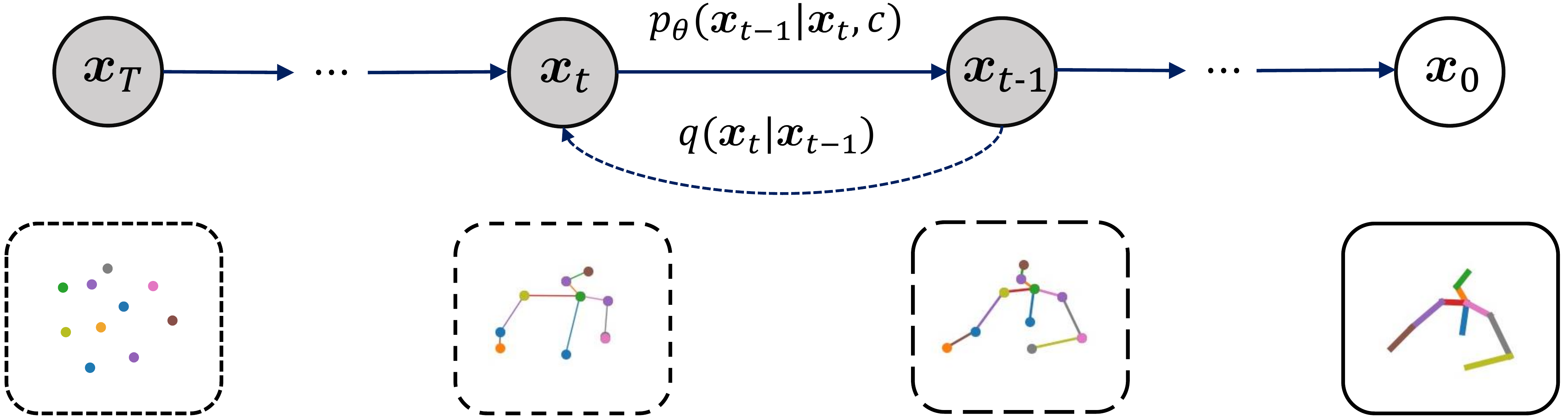}
\caption{\textbf{Illustration of Conditional Generation Process in Co-Speech Gesture Generation.} The diffusion process $q$ gradually adds Gaussian noise to the gesture sequence (\textit{i.e.}, $\bm{x}_0$ sampled from the real data distribution). The generation process $p_{\theta}$ learns to denoise the white noise (\textit{i.e.}, $\bm{x}_T$ sampled from the normal distribution) conditioned on context information $\bm{c}$. Note that $\bm{x}_t$ denotes the corrupted gesture sequence at the $t$-th diffusion step.}
\label{overview}
\vspace{-0.2cm}
\end{figure}

The recent paradigm of diffusion probabilistic models provides a new perspective for realistic generation~\cite{ho2020denoising, song2021scorebased}, facilitating high-fidelity synthesis with desirable properties such as good distribution coverage and stable training compared to GANs.
However, it is non-trivial to adapt existing diffusion models for co-speech gesture generation. 
Most existing conditional diffusion models deal with \textit{static} data and conditions~\cite{Saharia2022Photorealistic, ramesh2022hierarchical} (\textit{e.g.}, the image-text pairs without temporal dimension), while co-speech gesture generation requires generating \textit{temporally coherent} gesture sequences conditioned on continual audio clips.
Further, the commonly used denoising strategy in existing diffusion models samples independently and identically distributed (\textit{i.i.d.}) noises in latent space to increase diversity. However, this strategy tends to introduce variation for each gesture frame and lead to temporal inconsistency in skeleton sequences. 
Therefore, how to generate high-fidelity co-speech gestures with strong audio correlations and temporal consistency is quite challenging within the diffusion paradigm.

To address the above challenges, we propose a tailored Diffusion Co-Speech Gesture framework to \textit{capture the cross-modal audio-gesture associations while maintaining temporal coherence} for high-fidelity audio-driven co-speech gesture generation, named \textbf{DiffGesture}.
As shown in Figure~\ref{overview}, we formulate our task as a diffusion-conditional generation process on clips of skeleton and audio, where the diffusion phase is defined by gradually adding noise to gesture sequence, and the generation phase is referred as a parameterized Markov chain with conditional context features of audio clips to denoise the corrupted gestures. 
As we treat the multi-frame gesture clip as the diffusion latent space, the skeletons can be efficiently synthesized in a non-autoregressive manner to bypass error accumulation.
To better attend to the sequential conditions from multiple modalities and enhance the temporal coherence, we then devise a novel \textit{Diffusion Audio-Gesture Transformer} architecture to model audio-gesture long-term temporal dependency.
Particularly, the per-frame skeleton and contextual features are concatenated in the aligned temporal dimension and embedded as individual input tokens to a Transformer block.
Further, to eliminate the temporal inconsistency caused by the naive denoising strategy in the inference stage, we thus propose a new \textit{Diffusion Gesture Stabilizer} module to gradually anneal down the noise discrepancy in the temporal dimension.
Finally, we incorporate implicit classifier-free guidance by jointly training the conditional and unconditional models, which allows us to trade off between the diversity and sample quality during inference.

Extensive experiments on two benchmark datasets show that our synthesized results are coherent with stronger audio correlations and outperform the state-of-the-arts with superior performance on co-speech gesture generation.
To summarize, our main contributions are three-fold: \textbf{1)} As an early attempt at taming diffusion models for co-speech gesture generation, we formally define the diffusion and denoising process in gesture space, which synthesizes audio-aligned gestures of high-fidelity. \textbf{2)} We devise the \textit{Diffusion Audio-Gesture Transformer} with implicit classifier-free diffusion guidance to better deal with the input conditional information from multiple sequential modalities. \textbf{3)} We propose the \textit{Diffusion Gesture Stabilizer} to eliminate temporal inconsistency with an annealed noise sampling strategy. 
\section{Related Work}\label{sec:related-work}

\noindent\textbf{Co-Speech Gesture Generation.} Synthesizing co-speech gestures is crucial for a variety of applications. Conventional studies resort to rule-based pipelines~\cite{cassell1994animated, huang2012robot, marsella2013virtual}, where linguistic experts pre-define the speech-gesture pairs and refine the transitions between different motions. Recent works exploit neural networks to learn the mapping from speech to gesture based on a large training corpus, where an off-the-shelf pose estimator is leveraged to label the online videos for pseudo annotations~\cite{yoon2019robots, ginosar2019learning, liu2022learning, ahuja2020no, yoon2020speech, qian2021speech}. 
Meanwhile, some works study the influence of input modality, verifying the connections between co-speech gesture and speech audio~\cite{li2021audio2gestures}, text transcript~\cite{ahuja2019language2pose}, speaking style~\cite{ahuja2020style}, and speaker identity~\cite{yoon2020speech}. 
To further improve the model's capacity, previous studies explore multiple architecture choices, including CNN~\cite{habibie2021learning}, RNN~\cite{yoon2019robots}, Transformer~\cite{bhattacharya2021text2gestures}, and VQ-VAE~\cite{yazdian2022gesturevec, liu2022audio}. Notably, several recent works are based on GANs to guarantee realistic results~\cite{ginosar2019learning, yoon2020speech, qian2021speech, liu2022learning}, which involve the adversarial training between the generator and the discriminator. However, the notorious mode collapse and unstable training of GANs prevent the high-fidelity gesture distribution learning conditioned on audio.

\begin{figure*}[t]
\centering
\includegraphics[width=1.00\textwidth]{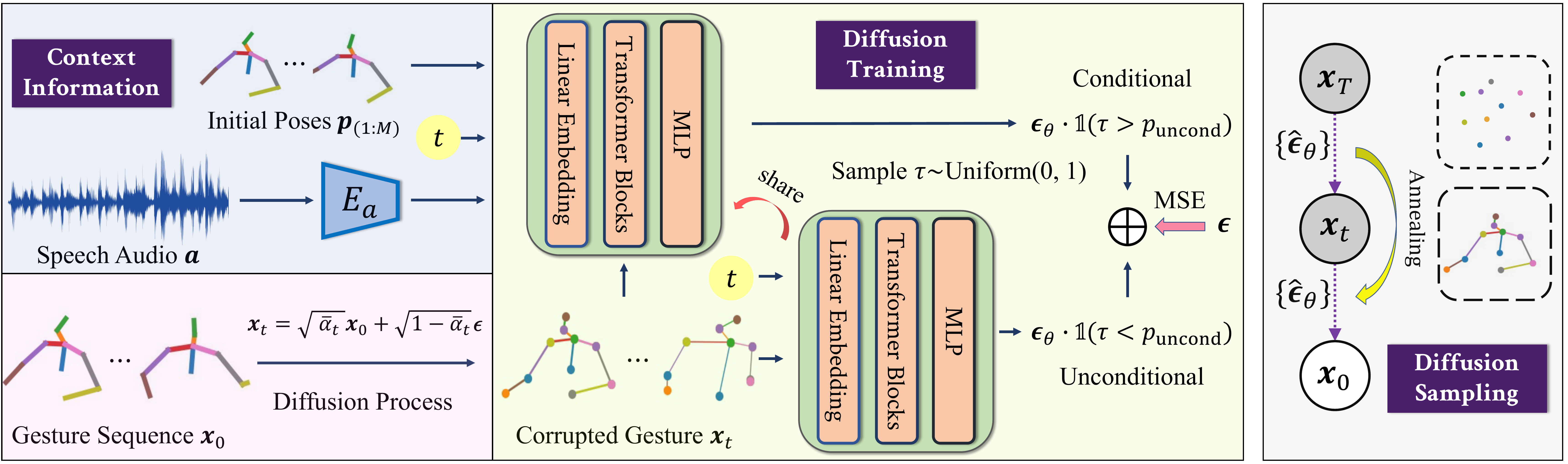}
\caption{\textbf{Overview of the Diffusion Co-Speech Gesture (DiffGesture) Framework.} Given the gesture sequence $\bm{x}_0$, we first establish the forward diffusion (\textcolor[rgb]{0.8, 0.2, 0.8}{purple}) and conditional denoising process (\textcolor[rgb]{0.3, 0.6, 0.2}{green}) in gesture space. Then, we devise the Diffusion Audio-Gesture Transformer to attend to the input conditions of initial poses $\bm{p}_{(1:M)}$, speech audio $\bm{a}$, time embedding $t$ and corrupted gesture $\bm{x}_t$ from multiple modalities (\textcolor[rgb]{0.18039215686275, 0.45882352941176, 0.71372549019608}{blue}). At the diffusion sampling stage (\textcolor[rgb]{0.37647058823529, 0.37647058823529, 0.37647058823529}{grey}), we propose the Diffusion Gesture Stabilizer to eliminate temporal inconsistency with an annealed noise sampling strategy. To further incorporate implicit classifier-free guidance, we jointly train the conditional ($1 - p_{\textit{uncond}}$) and unconditional ($p_{\textit{uncond}}$) models. This allows us to trade off between diversity and quality during inference.} 
\label{framework}
\vspace{-8pt}
\end{figure*}

\noindent\textbf{Diffusion Probabilistic Models.} Diffusion probabilistic models have achieved promising results on unconditional image generation~\cite{ho2020denoising}, which are further applied to conditional tasks like text-to-image~\cite{Saharia2022Photorealistic}. Among the diffusion-based literature, previous works focus on static data and conditions. Besides, they mainly utilize explicit guidance like pretrained classifiers~\cite{dhariwal2021diffusion} and CLIP similarity~\cite{nichol22glide, liu2023more, Saharia2022Photorealistic, ramesh2022hierarchical} to guide the generation process. In this work, we explore a more challenging co-speech gesture generation setting, where the gesture data and audio conditions are both sequential, and the audio-to-gesture mapping is implicit. To this end, we propose the Diffusion Audio-Gesture Transformer to guarantee temporally aligned generation. We further propose the Diffusion Gesture Stabilizer to simultaneously achieve diverse and temporally coherent gestures with an annealed noise sampling strategy.

\section{Our Approach}\label{sec:method}

Figure~\ref{framework} depicts an overview of the proposed \textbf{DiffGesture} framework to generate co-speech gestures of high fidelity.
In this section, we first introduce the problem formulation of audio-driven co-speech gesture generation (Section~\ref{sec:3.1}). We then establish the forward diffusion and the reverse conditional generation process in gesture space (Section~\ref{sec:3.2}). Furthermore, we elaborate the Diffusion Audio-Gesture Transformer to attend to the conditions from multiple modalities and enhance the speech-gesture correlations with temporal dependency (Section~\ref{sec:3.3}). To eliminate temporal inconsistency introduced by naive noises, we propose a novel Diffusion Gesture Stabilizer with annealed noise sampling strategies and describe this module in (Section~\ref{sec:3.4}). Finally, incorporating implicit classifier-free guidance in co-speech gestures is discussed in (Section~\ref{sec:3.5}).

\subsection{Problem Formulation}
\label{sec:3.1}
With a large-scale co-speech gesture training corpus, we leverage the speaking videos with clear co-speech upper body movements for model learning. In particular, for each video clip of $N$ frames, we extract the accompanying speech audio sequence $\bm{a} = \{\bm{a}_1, \dots, \bm{a}_N\}$ and use the off-the-shelf human pose estimator OpenPose~\cite{cao2019openpose} to annotate the per-frame human skeletons as $\bm{x} = \{\bm{p}_1, \dots, \bm{p}_N\}$. We follow baseline methods~\cite{yoon2020speech, liu2022learning} to further pre-process such skeletal representation into the concatenation of unit direction vectors as $\bm{p}_i = \left[\bm{d}_{i,1}, \bm{d}_{i,2}, \dots, \bm{d}_{i,J-1}\right]$, where $\bm{p}_i$ denotes the 2D keypoint coordinates of the $i$-th frame, $J$ is the total joint number and $\bm{d}_{i,j}$ represents the unit direction vector between the $j$-th and the ($j$+1)-th joint of the $i$-th image frame. The diffusion model's reverse denoising process $G$ parameterized by $\theta$ is optimized to synthesize the human skeleton sequence $\bm{x}$, which is further conditioned on the speech audio sequence $\bm{a}$ and the initial poses $\{\bm{p}_1, \dots, \bm{p}_M\}$ of the first $M$ frames. 
The learning objective of the overall framework can be formulated as
$\mathop{\arg\min}_{\theta} || \bm{x} - G_{\theta} ( \bm{a}, \bm{p}_1, \dots, \bm{p}_M) ||.$

\subsection{Gesture Space Forward and Reverse Process}
\label{sec:3.2}
Given $\bm{x}_0 \in \mathbb{R}^{N \times 3(J-1)}$ sampled from real data distribution $ q(\bm{x}_0)$, our goal is to learn a model distribution $p_{\theta}(\bm{x}_0)$ parameterized by $\theta$ that approximates $q(\bm{x}_0)$. Specifically, denoising diffusion probabilistic models (DDPMs)~\cite{ho2020denoising} define the latent variable models of the form $p_{\theta}(\bm{x}_0) = \int p_{\theta}(\bm{x}_{0:T}) d\bm{x}_{1:T}$, where $\bm{x}_{1:T}$ are latent variables in the same sample space as $\bm{x}_0$ with the same dimensionality.

\noindent\textbf{The Forward Diffusion Process.} The \textit{forward process}, which is also termed as the \textit{diffusion process}, approximates the posterior distribution $q(\bm{x}_{1:T}|\bm{x}_0)$. It is defined as a Markov chain that gradually adds Gaussian noise to the data sample according to a variance schedule $\beta_1, \dots, \beta_T$:
\begin{align} \label{eq:forward}
q(\bm{x}_{1:T} | \bm{x}_0) &= \prod_{t=1}^T q(\bm{x}_t | \bm{x}_{t-1}),\\
\textit{where }\quad q(\bm{x}_t | \bm{x}_{t-1}) &= \mathcal{N}(\bm{x}_t; \sqrt{1 - \beta_t}\bm{x}_{t-1}, \beta_t \bm{I}).
\end{align} 

The variances $\beta_t$ are constant hyperparameters to ease the modeling of the reverse process~\cite{ho2020denoising}. Through such a corruption scheme, the structural information of the original skeleton is gradually substituted by noises, which finally leads to a pure white noise when $T$ goes to infinity. Therefore, the prior latent distribution of $p(\bm{x}_T)$ is $\mathcal{N}(\bm{x}_T; \bm{0}, \bm{I})$ with only information of Gaussian noise.

\noindent\textbf{Reverse Conditional Gesture Generation.} The \textit{reverse process}, which is also termed as the \textit{generative process}, estimates the joint distribution of $p_{\theta}(\bm{x}_{0:T})$. As proved in~\cite{Feller1949OnTT}, the reverse process of the \textit{continuous} diffusion process preserves the same transition distribution form, which motivates us to leverage a Gaussian transition to formulate $p_{\theta}(\bm{x}_{t-1} | \bm{x}_{t})$ under an unconditional setting, which approximates the intractable process as:
\begin{align} 
\label{eq:reverse}
p_{\theta}(\bm{x}_{0:T}) &= p_{\theta}(\bm{x}_T) \prod_{t=1}^T p_{\theta}(\bm{x}_{t-1} | \bm{x}_{t}),\\
\textit{where }p_{\theta}(\bm{x}_{t-1}|\bm{x}_{t}) &= \mathcal{N}(\bm{x}_{t-1}; \mu_{\theta}(\bm{x}_t, t), \Sigma_{\theta}(\bm{x}_t, t)).
\label{eq:reverse2}
\end{align} 
The corrupted noisy data $\bm{x}_t$ is sampled by $q(\bm{x}_t | \bm{x}_0) = \mathcal{N}(\bm{x}_t; \sqrt{\bar{\alpha}_t}\bm{x}_0, (1 - \bar{\alpha}_t) \bm{I})$, where $\alpha_t = 1 - \beta_t$ and $\bar{\alpha}_t = \prod_{s=1}^t \alpha_s$. Note that we set the variances $\Sigma_{\theta}(\bm{x}_t, t)=\beta_t \bm{I}$ to untrained time-dependent constants.
 The above diffusion model formulations show compelling performances on unconditional generation. To further adapt to the conditional co-gesture synthesis, we have to provide additional inputs to the model, including the audio and initial poses. Therefore, we treat the speech audio $\bm{a}$ and initial poses $\bm{p}_{1:M}$ as context information $\bm{c}$ and inject conditions into the generation process. The reverse process of each timestep (Eq.~\ref{eq:reverse2}) can be thus updated as:
\begin{align} \label{eq:conditional}
p_{\theta}(\bm{x}_{t-1}|\bm{x}_{t}, \bm{c}) &= \mathcal{N}(\bm{x}_{t-1}; \mu_{\theta}(\bm{x}_t, t, \bm{c}), \beta_t \bm{I}).
\end{align} 
In this way, we could start the generation process by firstly sample a Gaussian noise $\bm{x}_T \sim \mathcal{N}(\bm{0}, \bm{I})$ and follow the Markov chain to iteratively denoise the latent variable $\bm{x}_t$ via Eq.~\ref{eq:conditional} to get the final results. The overview of conditional co-speech gesture process is illustrated in Figure~\ref{overview}.

\noindent\textbf{Training Objective.} To optimize the overall framework, we optimize the variational lower bound on negative log-likelihood: $\mathbb{E}[-\log p_{\theta}(\bm{x}_0)] \leq \mathbb{E}_{q}[-\log \frac{p_{\theta}(\bm{x}_0)}{q(\bm{x}_{1:T} | \bm{x}_0)}].$
We rewrite the loss function conditioned on context $\bm{c}$ and eliminate all the constant items that do not require training: 
$L(\theta) = \mathbb{E}_{q}[\sum_{t=2}^T D_{KL} (q(\bm{x}_{t-1} | \bm{x}_t, \bm{x}_0) || p_{\theta}(\bm{x}_{t-1} | \bm{x}_t, \bm{c}))]$.
With reparameterization, we can represent each term in $L_\theta$ using MSE loss. We follow~\cite{ho2020denoising} to further simplify the training objective to the ensemble of MSE losses as:

\begin{align}
    L (\theta) = \mathbb{E}_{q}[\left\| \bm{\epsilon} - \bm{\epsilon}_\theta(\sqrt{\bar\alpha_t} \bm{x}_{0} + \sqrt{1-\bar\alpha_t}\bm{\epsilon}, \bm{c}, t) \right\|^2],
\label{eq:losssimple}
\end{align}
where $t$ is uniformly sampled between 1 and $T$. 
As we jointly train the model under conditional and unconditional setting, a trainable masked embedding with probability $p_{\textit{uncond}}$ replaces context $\bm{c}$ and the diffusion model predicts the noise in the unconditional setting. The detailed principles will be discussed in Section~\ref{sec:3.5}.

\subsection{Diffusion Audio-Gesture Transformer}
\label{sec:3.3}
With the naive conditional generation scheme as specified in Section~\ref{sec:3.2}, we still confront a critical problem in the setting of co-speech gesture generation. Since $\bm{x}_0$ denotes the skeleton sequence of $N$ frames, there exists temporal dependency among the target sequence and context information, making it more complex than time-invariant tasks like image generation. Therefore, how to guarantee temporally coherent results in a non-autoregressive conditional generation process remains an unsolved problem.

In contrast to most previous studies that resort to recurrent networks~\cite{yoon2020speech, liu2022learning}, we propose to make use of the Transformer's strong capacity in sequential data modeling. Specifically, since the noisy gesture sequence $\bm{x}_t$ and the contextual information $\bm{c}$ align in the temporal dimension, we concatenate them in the feature channel. In this way, the skeleton and context condition of each frame serve as an individual token, which captures the long-term dependency by the self-attention mechanism:
\begin{align}
    \text{Attention}(\mathbf{Q, K, V}) = \text{softmax}(\frac{\mathbf{QK^\mathsf{T}}}{\sqrt{\ell}})\mathbf{V},
\label{eq:attention}
\end{align}
where $\mathbf{Q, K, V}$ are the query, key, and value matrix from input tokens, $\ell$ is the channel dimension, and $\mathsf{T}$ is the matrix transpose operation. Such a design also avoids severe error accumulation in autoregressive pipelines, enabling us to generate coherent gesture sequences.

\subsection{Diffusion Gesture Stabilizer}
\label{sec:3.4}

In DDPMs, the independent random variables $\bm{z}$ introduced at the sampling stage promote diversity and thus systematically improve the task performance. However, the variation in temporal dimension introduced by $\bm{z}$, especially when timestep $t$ is small in the reverse process, can have a negative effect on temporal consistency. At the inference stage, to achieve the trade-off between diversity and temporal consistency, we propose a novel Diffusion Gesture Stabilizer without extra training expenses under two \textbf{annealed} scenarios, where the term ``annealed'' means that the process is transitioned from high variance and entropy (hot) to low variance and entropy (cold). 

\noindent\textbf{Thresholding.} Since temporally independent Gaussian noises inevitably introduce inconsistency, restricting the temporal variation naturally helps to avoid inconsistency. And hard thresholding serves as an effective trick. In detail, we set a time threshold $t_0$, and then use the same $\bm{z} \in \mathbb{R}^{N \times C}$ in the naive sampling strategy for $t > t_0$ and set $\bm{z} = \{\bm{z}_0\}_{i=1}^{N}$ for $t \leq t_0$, where $z_0 \in \mathbb{R}^{C}$ follows $\mathcal{N}(\bm{0}, \bm{I})$ which do not introduce variation in the temporal dimension. 

\noindent\textbf{Smooth Sampling.} We further modify $\bm{z}(t)=\{\bm{z}_i(t)\}_{i=1}^{N}$ to be a smooth annealing version via variance-aware sampling. In the original sampling rule of DDPMs, \textit{i.i.d.}variables $\bm{z}_i(t)$ are sampled from $\mathcal{N}(\bm{0}, \bm{I})$. With smooth resampling, we first sample $\bm{z}_0(t) \sim \mathcal{N}(\bm{0}, \sigma^2_a(t)\bm{I})$ only once for each timestep $t$ in the reverse process, then given $\bm{z}_0(t)$, we sample $\bm{z}_i(t)|\bm{z}_0(t)\sim \mathcal{N}(\bm{z}_0(t), (1-\sigma^2_a(t))\bm{I})$ for $i \in \{1,\dots,N$\}, where $\sigma_a(t) \in 	\left[0,1\right]$ is a non-decreasing function to achieve variance annealing.

\algrenewcommand\algorithmicindent{1.0em}%
\begin{algorithm}[H]
  \caption{\textbf{Training}} \label{alg:training}
  \small
  \begin{algorithmic}[1]
    \Repeat
      \State Sample $(\bm{x}_{0},\bm{c}) \sim q(\bm{x}_{0},\bm{c})$
      \State Sample $\tau \sim \operatorname{Uniform}(0,1)$. Set $\bm{c} = \varnothing$  if $\tau < p_{\textit{uncond}}$
      \State Sample $t \sim \operatorname{Uniform}(\{ 1, \ldots, T\})$
      \State Compute $\nabla_{\theta} \left\| \bm{\epsilon} - \bm{\epsilon}_\theta(\sqrt{\bar\alpha_t} \bm{x}_{0} + \sqrt{1-\bar\alpha_t}\bm{\epsilon}, \bm{c}, t) \right\|^2$
      \State Perform gradient descent
    \Until{converged}
  \end{algorithmic}
\end{algorithm}

\subsection{Implicit Classifier-free Guidance}
\label{sec:3.5}
In co-speech gesture literature, the speech-to-gesture mapping is implicit, where the same audio corresponds to diverse gestures and different audios could incur the same motion~\cite{lee2021crossmodal, li2021audio2gestures}, making it difficult to utilize the commonly used explicit classifier guidance~\cite{dhariwal2021diffusion, nichol22glide}. Therefore, 
how can we further exploit practical guidance for better audio correlations and mode coverage? Our solution to this question is to train an extra ``unconditional'' diffusion model to implicitly guide the generation. Dhariwal \textit{et al.}~\cite{dhariwal2021diffusion} first introduce the classifier guidance of cross-entropy gradient $\nabla_{\bm{x}_t}\log p_{\phi}(y|\bm{x}_t)$, where the pretrained classifier is parameterized by $\phi$ and $y$ denotes the classification logits. This gradient term is further scaled by the covariance matrix to modify the mean value of transition distribution in Eq.~\ref{eq:reverse2}. To adapt to the cases where no explicit guidance is available, we follow~\cite{ho2021classifierfree} to jointly train the conditional and unconditional models, termed as classifier-free guidance. In particular, according to the implicit classifier's property that $p^i(y|\bm{x}_t) \propto p(\bm{x}_t | y) / p(\bm{x}_t)$, we could derive a gradient relationship in the implicit classifier as:
\begin{align}
    \nabla_{\bm{x}_t}\log p^i(y|\bm{x}_t) \propto \nabla_{\bm{x}_t}\log p(\bm{x}_t|y) - \nabla_{\bm{x}_t}\log p(\bm{x}_t),
\label{eq:guidance}
\end{align}
which is further proportional to $\bm{\epsilon}^*(\bm{x}_t | y) - \bm{\epsilon}^*(\bm{x}_t)$. Therefore, we use a single Transformer network to parameterize both settings by a mix-up training trick: for the probability of $p_{uncond}$, we set the context information $\bm{c}$ as masked embedding to train the unconditional setting, while for other cases, we train the original conditional counterpart. The training is shown in Algorithm~\ref{alg:training}.

\noindent\textbf{Sampling with Classifier-free Guidance.} Starting from Gaussian noise, we iteratively remove noises in $x_t$. As we use implicit classifier-free guidance, similar to Equation~\ref{eq:guidance}, the predicted Gaussian noise is modified as:
\begin{align}
   \hat{\bm{\epsilon}}_\theta = \bm{\epsilon}_\theta(\bm{x}_t, t) + s \cdot (\bm{\epsilon}_{\theta}(\bm{x}_t, \bm{c}, t) - \bm{\epsilon}_\theta(\bm{x}_t, t)),
\label{eq:epsmodified}
\end{align}
where $s$ is the scale parameter to trade off the diversity and quality. With classifier-free guidance, Algorithm~\ref{alg:sampling} reveals how to generate co-speech gestures given the trained diffusion model via the Diffusion Gesture Stabilizer with the Smooth Sampling annealed scenario.

\algrenewcommand\algorithmicindent{1.0em}%
\begin{algorithm}[H]
  \caption{\textbf{Sampling}} \label{alg:sampling}
  \small
\begin{algorithmic}[1]
     \State Trained diffusion model $\theta$, $\bm{x}_T \sim \mathcal{N}(\bm{0}, \bm{I})$
    \For{$t=T, \dotsc, 1$}
      \State $\hat{\bm{\epsilon}}_\theta = \bm{\epsilon}_\theta(\bm{x}_t, t) + s \cdot (\bm{\epsilon}_{\theta}(\bm{x}_t, \bm{c}, t) - \bm{\epsilon}_\theta(\bm{x}_t, t))$
        \State $\bm{z}_0(t) \sim \mathcal{N}(\bm{0}, \sigma^2_a(t)  \bm{I})$ 
        \For{$i=1, \dotsc, N$}
        \State $\bm{z}_i(t) \sim \mathcal{N}(\bm{z}_0(t) , (1-\sigma^2_a(t)) \bm{I})$ 
    \EndFor
      \State $\bm{z}(t)=\{\bm{z}_1(t),\dots,\bm{z}_N(t)\}$, if $t > 1$, else $\bm{z}(t) = \bm{0}$
      \State $\bm{x_{t-1}} = \frac{1}{\sqrt{\alpha_t}}\left( \bm{x_t} - \frac{1-\alpha_t}{\sqrt{1-\bar\alpha_t}} \hat{\bm{\epsilon}}_\theta \right) + \sigma_t \bm{z}(t)$
    \EndFor
    \State \textbf{return} $\bm{x}_0$
  \end{algorithmic}
\end{algorithm}
\section{Experiments}\label{sec:exp}
\begin{table*}[ht]
  \centering
  \begin{tabular}{lcccccccc}
    \toprule
     & \multicolumn{3}{c}{TED Gesture~\cite{yoon2020speech}} &  \multicolumn{3}{c}{TED Expressive~\cite{liu2022learning}} \\
    \cmidrule(r){2-4} \cmidrule(r){5-7}
    Methods & FGD $\downarrow$ & BC $\uparrow$ & Diversity $\uparrow$ & FGD $\downarrow$ & BC $\uparrow$ & Diversity $\uparrow$ \\
    \midrule
    Ground Truth & 0 & 0.698 & 108.525 & 0 & 0.703 & 178.827\\
    \midrule
    Attention Seq2Seq~\cite{yoon2019robots}   & 18.154  & 0.196 & 82.776 & 54.920 & 0.152 & 122.693\\
     Speech2Gesture~\cite{ginosar2019learning}  & 19.254  & 0.668 & 93.802 & 54.650 & 0.679 & 142.489\\
     Joint Embedding~\cite{ahuja2019language2pose}   & 22.083  & 0.200 & 90.138 & 64.555 & 0.130 & 120.627\\
     Trimodal~\cite{yoon2020speech}   & 3.729 & 0.667 & 101.247 & 12.613 & 0.563 & 154.088\\
    HA2G~\cite{liu2022learning} & 3.072 & 0.672 & 104.322 & 5.306 & 0.641 & 173.899\\
     \midrule 
    \textbf{DiffGesture (Ours)} & \textbf{1.506} & \textbf{0.699} & \textbf{106.722} &      \textbf{2.600} & \textbf{0.718} & \textbf{182.757} \\

    \bottomrule
  \end{tabular}
  \caption{\textbf{The Quantitative Results on TED Gesture~}\cite{yoon2020speech} \textbf{and TED Expressive~}\cite{liu2022learning}. We compare the proposed diffusion-based method against recent SOTA methods~\cite{ahuja2019language2pose, ginosar2019learning, yoon2020speech, yoon2019robots, liu2022learning} and ground truth. For FGD, the lower, the better; for other metrics, the higher, the better.}
  \label{tbl:res}
\end{table*}

\subsection{Co-Speech Gesture Datasets}

\noindent \textbf{TED Gesture.} As a large-scale dataset for gesture generation research, TED Gesture dataset~\cite{yoon2019robots, yoon2020speech} contains 1,766 TED videos of different narrators covering various topics. 
We follow the data process in former works~\cite{yoon2020speech, liu2022learning}, where the poses are resampled with 15 FPS, and frame segments of length 34 are obtained with a stride of 10.

\noindent 
\textbf{TED Expressive.} While the poses in TED Gesture only contain 10 upper body key points without vivid finger movements, the TED Expressive dataset~\cite{liu2022learning} is further expressive of both finger and body movements. The state-of-art 3D pose estimator ExPose~\cite{ExPose:2020} is used to fully capture the pose information in data. As a result, TED Expressive annotates the 3D coordinates of 43 keypoints, including 13 upper body joints and 30 finger joints.

\subsection{Experimental Settings}

\noindent 
\textbf{Comparison Methods.} 
We compare our method on two benchmark datasets with the state-of-the-art methods in recent years. \textbf{1) Attention Seq2Seq}~\cite{yoon2019robots} elaborates on the attention mechanism to generate pose sequences from speech text. 
\textbf{2) Speech2Gesture}~\cite{ginosar2019learning} uses spectrums of the speech audio segments as the input and generates speech gestures adversarially. 
\textbf{3) Joint Embedding}~\cite{ahuja2019language2pose} maps text and motion to the same embedding space, then generates outputs from motion description text.
\textbf{4) Trimodal}~\cite{yoon2020speech} serves as a strong baseline that learns from text, audio, and speaker identity to generate gestures, outperforming former methods by a large margin.
\textbf{5) HA2G}~\cite{liu2022learning} introduces a hierarchical audio learner that captures information across different semantic granularities, achieving state-of-the-art performances. This method hierarchically extracts rich features at the cost of heavier GPU memory overhead, while our method requires much smaller expenses.

\noindent 
\textbf{Implementation Details.} For all the methods in both datasets, we set $N=34$ and $M=4$ to get $M$-frame pose sequences where the first $N$ frames are used for reference, termed as initial poses. There are $J$ upper body joints in all the frames of pose sequences, where $J=10$ for TED Gesture and $J=43$ for TED Expressive. Following~\cite{yoon2020speech}, to eliminate the effect of the joint lengths and root motion, we represent the joints' positions using $J-1$ directional vectors normalized to the unit vectors and train the model to learn the directional vectors.
For the audio processing, we use the same audio encoder used in~\cite{yoon2020speech} to extract the feature of the raw audio clips directly. 
The audio clips are encoded as $N$ audio feature vectors of 32-D.
The audio feature and initial poses are concatenated to form the conditional context information of the diffusion model. For the diffusion process, the number of timesteps is $T=500$, and the variances increase linearly from $\beta_1=1e-4$ to $\beta_T=0.02$.
For the Stabilizer, $t_0$ can be adjusted from 20-30 for Thresholding, and a quadratic non-increasing function $\sigma_a(t)$ is applied for Smooth Sampling. The hidden dimension of the transformer blocks, is set to 256 for TED Gesture and 512 for TED Expressive. We use 8 Transformer blocks, each of which comprises a multi-head self-attention block and a Feed-Forward Network. We use an Adam optimizer, and the learning rate is $5e-4$. It takes 10 hours to train the model on TED Gesture and 20 hours on TED Expressive on a single NVIDIA GeForce RTX 3090 GPU.

\begin{figure*}[t]
\centering
\includegraphics[width=1.00\textwidth, height=0.59\textwidth]{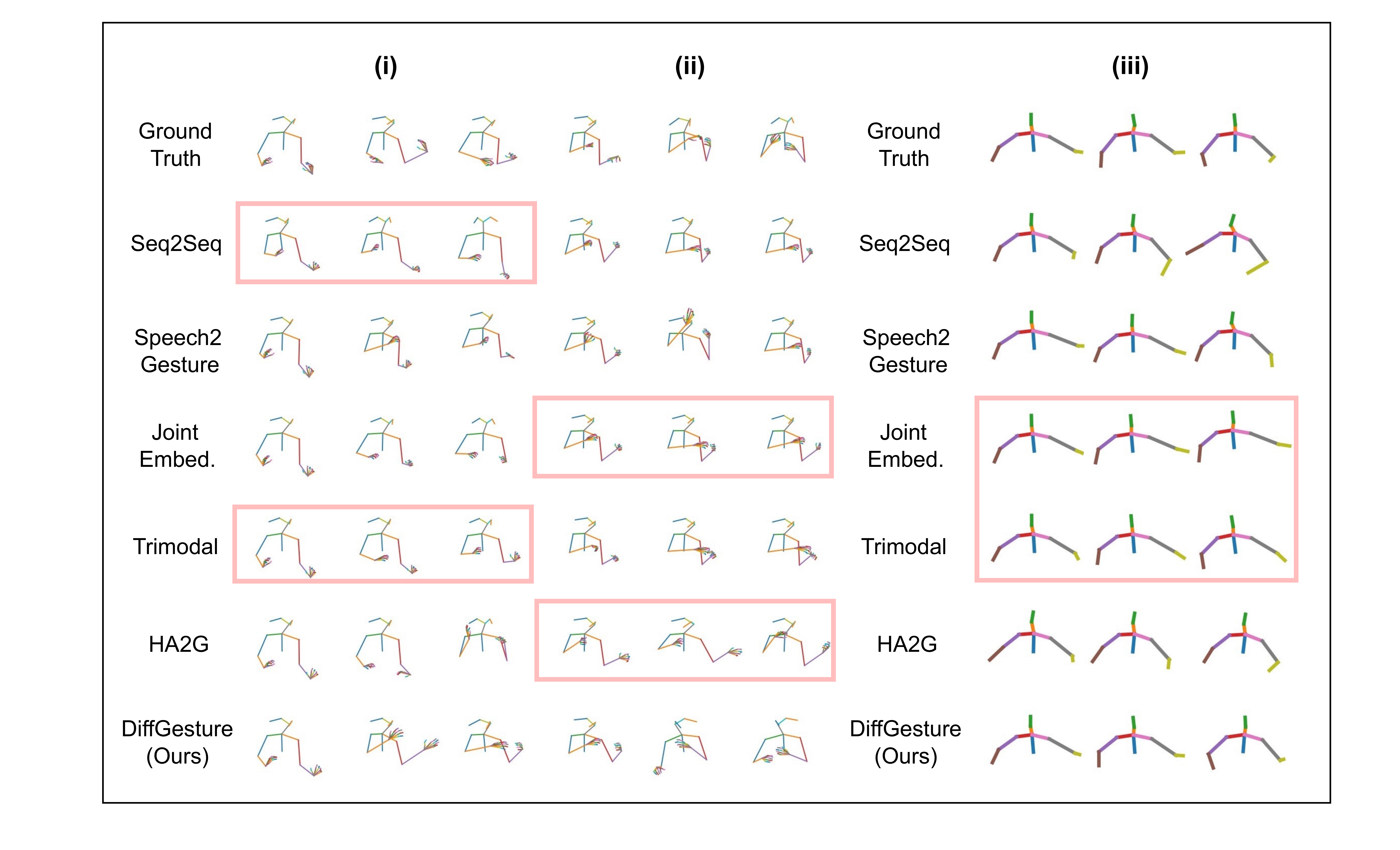}
\caption{\textbf{Visualization Results of Our DiffGesture on Two Datasets.} Three cases are picked up, where (i) and (ii) are TED Expressive cases, and (iii) is a TED Gesture case. We highlight dull cases generated by comparison methods with rectangles, indicating the mode collapse phenomenon of baselines.}
\label{result}
\end{figure*}

\begin{table*}
  \centering
  \begin{tabular}{cccccccc}
    \toprule
    Methods &
    GT &
    S2S.~\cite{yoon2019robots}& 
    S2G.~\cite{ginosar2019learning} & 
    Joint.~\cite{ahuja2019language2pose} & Tri.~\cite{yoon2020speech} & 
     HA2G~\cite{liu2022learning} & 
    \textbf{DiffGesture(Ours)}  \\
    \midrule
    Naturalness & 4.33 & 1.22 & 2.56 & 1.22 & 3.22 & 3.67 &\textbf{4.00}\\
    Smoothness & 3.94 & 3.50 & 1.61 & 3.44 & 3.44 & 3.39 & \textbf{3.89}\\
    Synchrony & 4.00 & 1.67 & 3.17 & 1.39 & 3.28 & 3.44 &\textbf{3.89}\\
    \bottomrule
  \end{tabular}
  \caption{\textbf{User Study Results.} The ratings of motion naturalness, smoothness, and synchrony, are on a scale of 1-5, with 5 being the best.}
  \label{table:userstudy}
\end{table*}

\subsection{Evaluation Metrics}
In evaluation, we use three metrics that are used in co-speech gesture generation and relative fields~\cite{liu2022learning, li2022danceformer}. 

\noindent\textbf{Fr\'echet Gesture Distance (FGD).} Similar to the Fr\'echet Inception Distance (FID) metric~\cite{heusel2017gans}, which is widely applied in image generation studies, FGD is used to measure the distance between the synthesized gesture distribution and the real data distribution. Yoon \textit{et al.}~\cite{yoon2020speech} define FGD by training a skeleton sequence auto-encoder to extract the features of the real gesture sequences $X$ and the features of the generated gesture sequences $\hat{X}$:
\begin{equation} 
    \label{eq:fgd}
    \mathrm{FGD}(X,\hat{X}) = \|\mu_r-\mu_g\|^2+\rm{Tr}(\Sigma_r+\Sigma_g-2(\Sigma_r\Sigma_g)^{1/2}), 
    \nonumber
\end{equation}
where $\mu_r$ and $\Sigma_r$ are the first and the second moments of the latent feature distribution of the real gestures $X$, and $\mu_g$ and $\Sigma_g$ are the first and the second moments of the latent feature distribution of the generated gestures $\hat{X}$. 
We intuitively find that among the three metrics, FGD tells the most whether the generated pose sequences are of high quality.

\begin{table*}
  \centering
  \begin{tabular}{lcccccccc}
    \toprule
     & \multicolumn{3}{c}{TED Gesture~\cite{yoon2020speech}} &  \multicolumn{3}{c}{TED Expressive~\cite{liu2022learning}} \\
    \cmidrule(r){2-4} \cmidrule(r){5-7}
    Methods & FGD $\downarrow$ & BC $\uparrow$ & Diversity $\uparrow$ & FGD $\downarrow$ & BC $\uparrow$ & Diversity $\uparrow$ \\
    \midrule
    DiffGesture Base & 2.450 & 0.632 & 104.688 & 3.822 & 0.707 & 174.377\\
    DiffGesture w/o Stabilizer  & 2.219 & 0.674 & 105.192 & 2.792 & \textbf{0.721} & 180.125 \\
    DiffGesture w/o classifier-free & 1.810 & 0.673 & 105.644 & 3.326 & 0.717 & 178.245 \\
    \textbf{DiffGesture (Ours)} & \textbf{1.506} & \textbf{0.699} & \textbf{106.722} &      \textbf{2.600} & 0.718 & \textbf{182.757} \\
    \bottomrule
  \end{tabular}
  \caption{\textbf{Ablation Study on the Proposed Modules.} We investigate effectiveness of the proposed modules, Diffusion Gesture Stabilizer and implicit classifier-free guidance. The results indicate that the proposed modules consistently improve performance on the benchmarks.}
  \label{tbl:abl}
\end{table*}

\noindent \textbf{Beat Consistency Score (BC).} Proposed in~\cite{li2021learn, li2022danceformer}, BC measures motion-audio beat correlation. Considering that the kinematic velocities vary from different joints, we use the change of included angle between bones to track motion beats following~\cite{liu2022learning}. Specifically, we can calculate the mean absolute angle change (MAAC) of angle $\theta_j$ in adjacent frames by $\mathrm{MAAC}(\theta_j) = \frac{1}{S} \frac{1}{T-1} \sum_{s=1}^S\sum_{t=1}^{T-1}\|\theta_{j, s, t+1} - \theta_{j, s, t}\|_1$, where $S$ denotes the total number of clips in the dataset, $T$ denotes the number of frames in each clip, and $\theta_{j, s, t}$ is the included angle between the $j$-th and the ($j$+1)-th bone of the $s$-th clip at time-step $t$. Then, we can compute the angle change rate of frame $t$ for the $s$-th clip as $\frac{1}{J-1}\sum_{j=1}^{J-1}(\|\theta_{j, s, t+1} - \theta_{j, s, t}\|_1/\mathrm{MAAC}(\theta_j))$. Then we extract the local optima whose first-order difference is higher than a threshold to get kinematic beats, which 
are used to compute BC later. Following~\cite{li2022danceformer} to detect audio beat by onset strength~\cite{ellis2007beat}, we compute the average distance between each audio beat and its nearest motion beat as Beat Consistency Score:
\begin{equation} 
    \label{eq:bc}
    \mathrm{BC} = \frac{1}{n}\sum_{i=1}^n\exp (-\frac{\min_{\forall  t_j^y\in B^y}\|t_i^x - t_j^y\|^2}{2\sigma^2}),
\end{equation}
where $t^x_i$ is the $i$-th audio beats, $B^y=\{t^y_i\}$ is the set of the kinematic beats, and $\sigma$ is a parameter to normalize sequences, set to $0.1$ empirically.

\noindent\textbf{Diversity.} This metric evaluates the variations among generated gestures corresponding to various inputs~\cite{NEURIPS2019_7ca57a9f}. We use the same feature extractor when measuring FGD to map synthesized gestures into latent feature vectors and calculate the mean feature distance. In detail, we randomly pick 500 generated samples and compute the mean absolute error between the features and the shuffled features.

\subsection{Evaluation Results}

\noindent \textbf{Quantitative Results.} We compare our method with all the baselines with three metrics on TED Gesture and TED Expressive. The results are shown in Table~\ref{tbl:res}. For the metrics of Ground Truth, we report the values in our implementation. For TED Gesture, we report FGD of all baselines in~\cite{liu2022learning} and evaluate BC and Diversity on our own\footnote[2]{Since there exists an evaluation bug for the BC metric in HA2G~\cite{liu2022learning}, we report the re-implemented results from Liu \textit{et al.}}. For TED Expressive, all the results of baselines are reported from~\cite{liu2022learning}. Assuming the pseudo ground truth pose follows the real distribution, the FGD of Ground Truth in the table is 0. It is observed that our \textbf{DiffGesture} achieves state-of-the-art performance on both datasets, especially outperforming existing methods by a large margin on TED Expressive. Besides, since BC and Diversity are proposed to measure motion-audio beat correlation and variation, these two metrics of Ground Truth cannot be treated as upper bounds, and it is worth noting that our results may be higher than the ones of Ground Truth, indicating that the generated gestures are of high quality.    
\noindent \textbf{Qualitative Results.} We show the keyframes of all the methods on two datasets in Figure~\ref{result}. Since TED Expressive requires a higher ability of generative models, we pick up two cases for TED Expressive and one for TED Gesture. For each case, we select three keyframes (an early, a middle, and a late one) to show the pose motions. Comparison methods tend to generate slow and invariant poses and sometimes produce unreliable and stiff results. In contrast, DiffGesture produces diverse human-like poses without resulting in mean poses which are slow and rigid. Besides, as a primary drawback of GAN-based methods, mode collapse makes comparison methods often produce a single type of output, which is severe for pose motion generation. Such a phenomenon is shown in Fig.~\ref{result}, where the generated frames with nearly the same pose are highlighted with rectangles. 

\noindent \textbf{User Study.} To better validate the qualitative performance, we conduct a user study on the generated co-speech gestures. The study involves 18 participants, with 9 females and 9 males in the age range of 18-25 years old. The participants are required to grade the motion's quality and coherence, and all the clips are without labels. In total, we pick up 30 cases, 20 for TED-Expressive and 10 for TED Gesture. For each case, we show 7 videos with the order of the methods shuffling, including the ground truth. We adopt the Mean Opinion Scores rating protocol, and each participant is required to rate three aspects of generated motions: \textit{Naturalness}; \textit{Smoothness}; \textit{Synchrony between speech and generated gestures}. The results are shown in Table~\ref{table:userstudy} where the ratings are on a scale of 1 to 5, with 5 being the best. Our participants widely accept that our method produces high-fidelity results in all three aspects.

\subsection{Ablation Studies}
\noindent \textbf{Ablation Study on the Proposed Modules.} 
To demonstrate the effectiveness of our proposed \textbf{DiffGesture}, we present ablation studies on the key modules in the framework. In detail, we conduct experiments as follows. \textbf{1)} DiffGesture Base means we only use the proposed conditional diffusion generation process without further design.
\textbf{2)} DiffGesture w/o classifier-free, where classifier-free guidance is not implemented at both the training stage and inference stage. \textbf{3)} DiffGesture w/o Stabilizer, where Diffusion Gesture Stabilizer is removed at the inference stage. The results are reported in Table~\ref{tbl:abl}. The results illustrate the effectiveness of the designed Diffusion Gesture Stabilizer and the implicit classifier-free guidance.

\begin{table}
  \centering
  \begin{tabular}{lccccc}
    \toprule
    Methods & FGD $\downarrow$ & BC $\uparrow$ & Diversity $\uparrow$ \\
    \midrule
    GRU on $D_a$ & 14.343  & 0.658 & 98.472 \\
    Transformer on $D_a$& \textbf{1.506} & \textbf{0.699} & \textbf{106.722} \\
        \midrule
    GRU on $D_b$ & 17.452  & 0.680 & 172.168  \\
    Transformer on $D_b$& \textbf{2.600} & \textbf{0.718} & \textbf{182.757} \\
    \bottomrule
  \end{tabular}
  \caption{\textbf{Ablation Study on the Network Architectures.} We compare the performance of GRU and Transformer for diffusion-based backbone on TED Gesture ($D_a$) and TED Expressive ($D_b$).}
  \vspace{-1pt}
  \label{tbl:abl2}
\end{table}

\noindent \textbf{Ablation Study on the Network Architectures.} 
We investigate the performance of the GRU architecture in diffusion models, autoregressively generating poses in~\cite{yoon2020speech, liu2022learning}. We replace the Diffusion Audio-Gesture Transformer with the GRU in the diffusion model. All the context inputs remain the same as our method and are concatenated before the diffusion network. Results are shown in Table~\ref{tbl:abl2}. Though GRU serves as a strong baseline network in co-gesture learning, it fails to generate high-performance data like our designed Transformer-based network, which indicates the effectiveness of our Transformer-based network and that applying diffusion models in the audio-driven conditional generation is a non-trivial task.

\section{Conclusion}
\label{sec:conclusion}

In this work, we present a novel diffusion-based framework \textbf{DiffGesture} for co-speech gesture generation. To generate coherent gestures with strong audio correlations, we propose the Diffusion Audio-Gesture Transformer with the Diffusion Gesture Stabilizer to better attend to the condition information. Such a non-autoregressive pipeline helps to efficiently generate results and reduce error accumulation. We hope our method offers a new perspective for diffusion-based temporal generation and how to capture sequential cross-modal dependencies.

\noindent \textbf{Acknowledgement.}
The work described in this paper was partially supported by grants from the Research Grants Council of the Hong Kong Special Administrative Region, China (T45-401/22-N), the National Natural Science Fund (62201483), HKU Seed Fund for Basic Research (202009185079 and 202111159073), RIE2020 Industry Alignment Fund – Industry Collaboration Projects (IAF-ICP) Funding Initiative, as well as cash and in-kind contribution from the industry partner(s).
{\small

\bibliographystyle{IEEEtran}
}

\end{document}